\documentclass[sigconf]{aamas} 

\usepackage{balance} 
\usepackage{graphicx}
\usepackage{subcaption}
\usepackage{booktabs} 
\usepackage{comment}
\usepackage{hhline} 
\usepackage{environ} 

\usepackage{amsmath}
\usepackage{amsfonts}
\usepackage{algpseudocode,algorithm}
\algrenewcommand\algorithmicrequire{\textbf{Input:}}
\algrenewcommand\algorithmicensure{\textbf{Output:}}
\usepackage{xcolor}
\usepackage{bm}
\usepackage{tikz}
\usepackage{pgfplots}
\usepackage{pgfplotstable}

\usetikzlibrary{automata,positioning,arrows.meta,math,external,shapes.geometric,decorations.pathreplacing,calligraphy,fadings,matrix}
\pgfplotsset{compat=1.16}
\usepgfplotslibrary{fillbetween}

\setcopyright{ifaamas}
\acmConference[AAMAS '22]{Proc.\@ of the 21st International Conference
on Autonomous Agents and Multiagent Systems (AAMAS 2022)}{May 9--13, 2022}
{Online}{P.~Faliszewski, V.~Mascardi, C.~Pelachaud,
M.E.~Taylor (eds.)}
\copyrightyear{2022}
\acmYear{2022}
\acmDOI{}
\acmPrice{}
\acmISBN{}

\acmSubmissionID{325}

\title[Pareto Conditioned Networks]{Pareto Conditioned Networks}

\author{Mathieu Reymond}
\affiliation{
  \institution{Vrije Universiteit Brussel}
  \city{Brussels}
  \country{Belgium}}
\email{mreymond@ai.vub.ac.be}

\author{Eugenio Bargiacchi}
\affiliation{
  \institution{Vrije Universiteit Brussel}
  \city{Brussels}
  \country{Belgium}}
\email{ebargiac@ai.vub.ac.be}

\author{Ann Now\'{e}}
\affiliation{
  \institution{Vrije Universiteit Brussel}
  \city{Brussels}
  \country{Belgium}}
\email{ann.nowe@ai.vub.ac.be}

\begin{abstract}
In multi-objective optimization, learning all the policies that reach Pareto-efficient solutions is an expensive process. The set of optimal policies can grow exponentially with the number of objectives, and recovering all solutions requires an exhaustive exploration of the entire state space. We propose Pareto Conditioned Networks (PCN), a method that uses a single neural network to encompass all non-dominated policies. PCN associates every past transition with its episode's return. It trains the network such that, when conditioned on this same return, it should reenact said transition. In doing so we transform the optimization problem into a classification problem. We recover a concrete policy by conditioning the network on the desired Pareto-efficient solution. Our method is stable as it learns in a supervised fashion, thus avoiding moving target issues. Moreover, by using a single network, PCN scales efficiently with the number of objectives. Finally, it makes minimal assumptions on the shape of the Pareto front, which makes it suitable to a wider range of problems than previous state-of-the-art multi-objective reinforcement learning algorithms.
\end{abstract}


\newcommand{\BibTeX}{\rm B\kern-.05em{\sc i\kern-.025em b}\kern-.08em\TeX}

\providecommand{\method}{PCN }
\newcommand{\norm}[1]{\left\lVert#1\right\rVert}

\pgfplotsset{
    xjitter/.style={
        x filter/.code={\pgfmathparse{\pgfmathresult+rnd*#1-#1/2}}
    },
    xjitter/.default=0.1
}

\pgfplotsset{
    yjitter/.style={
        y filter/.code={\pgfmathparse{\pgfmathresult+rnd*#1}}
    },
    yjitter/.default=0.1
}

\DeclareMathOperator*{\argsort}{arg\,sort}

\makeatletter
\newsavebox{\measure@tikzpicture}
\NewEnviron{scaletikzpicturetowidth}[1]{%
  \def\tikz@width{#1}%
  \def\tikzscale{1}\begin{lrbox}{\measure@tikzpicture}%
  \BODY
  \end{lrbox}%
  \pgfmathparse{#1/\wd\measure@tikzpicture}%
  \edef\tikzscale{\pgfmathresult}%
  \BODY
}
\makeatother

\begin{document}


\pagestyle{fancy}
\fancyhead{}


\maketitle 


\section{Introduction}
\label{sec:introduction}

Decision makers acting in real-world problems often have to take into account multiple objectives. When maximizing one objective comes at the cost of another, the objectives are in conflict, and the decision maker must find a compromise between them. For example, maximizing the electricity output of a hydroelectric power plant comes at the expense of increased flooding risks downstream as well as irrigation deficits \cite{castelletti2013multiobjective}. In the medical field, radiotherapy should generally maximize the destruction of cancer cells while minimizing the damage to the healthy surrounding tissue\cite{jalalimanesh2017multi}. The optimal trade-offs might differ on a case-by-case basis, and they can be difficult to identify without a picture of all possible options.

Unfortunately, research work on optimizing a sequential decision problem often focuses on maximizing a single, scalar objective. For the aforementioned real-world problems, this means somehow combining the different objectives into a single target metric. A possible option is for the decision maker to ask an expert to manually engineer a metric so to induce a specific behavior after optimization. However, this approach has several drawbacks. First, such work can be expensive and error-prone, as it often requires domain expertise and extensive tuning before the optimization process finally produces a behavior that satisfies the decision maker. By focusing on a single solution at a time, this process also fails to inform the decision maker of all the trade-offs that could be possible. Finally, the decision maker has no way to convey directly their actual preferences, instead having to rely on the expertise of the designer to construct a suitable metric. These issues can severely restrict the influence that the decision maker has on the process, thus possibly missing their preferred solutions.

In contrast we can directly learn the best compromises by using an explicitly multi-objective approach. Assuming that improving an objective is always preferable, we can build a set of all optimal trade-offs called a \emph{Pareto front}. Once learned, the Pareto front stays fixed, since it does not depend on the preferences of the decision maker. The decision maker can then use the Pareto front to review all available policies, and use this knowledge to select their preferred one \cite{roijers2013}. This greatly simplifies the task of the decision maker, as the consequences of any given choice are clear and explainable in advance. Additionally, if the decision maker changes their mind at a later time, they can simply select a different policy without the need to re-tune an optimization metric and perform the learning process again.

In this work we propose a novel method, \emph{Pareto Conditioned Networks} (PCN), that is able to efficiently learn the policies that belong to the Pareto front. \method conditions a single neural network on the desired compromise, so that it outputs the policy predicted to achieve it. Our method is sample-efficient, as it feeds the experience used for different compromises into the same network, thus allowing to share experience across policies. This also means it does not need to learn a policy independently for each trade-off found in the Pareto front, which is an approach taken by several works in multi-objective reinforcement learning (MORL) \cite{roijers2015,parisi2014}. Moreover, we make minimal assumptions concerning the utility function --- i.e., the range of possible preferences of the decision maker --- as opposed to the often-used assumption of linear scalarization of the objectives in the MORL literature \cite{abels2019,runzhe2019}. Finally, our method is scalable with respect to the number of objectives, as opposed to many of the current state-of-the-art MORL methods, who often limit themselves to 2- or 3-objective problems \cite{hayes2021}.

\section{Background}
\label{sec:background}

\subsection{Multi-Objective Reinforcement Learning}
\label{sec:morl}

In \emph{reinforcement learning (RL)}, an agent learns to optimise its behaviour by interacting with the environment. In this paper, we deal with decision problems with multiple objectives, and model this as a \emph{multi-objective Markov decision process (MOMDP)}. A MOMDP is a tuple, $\mathcal{M} = \langle \mathcal{S}, \mathcal{A}, \mathcal{T}, \gamma, \vec{\mathcal{R}}, n \rangle$, where $\mathcal{S}, \mathcal{A}$ are the state and action spaces respectively, $\mathcal{T} \colon \mathcal{S} \times \mathcal{A} \times \mathcal{S} \to \left[ 0, 1 \right]$ is a probabilistic transition function, $\gamma$ is a discount factor determining the importance of future rewards and $\vec{\mathcal{R}} \colon \mathcal{S} \times \mathcal{A} \times \mathcal{S} \to \mathbb{R}^n$ is an $n$-dimensional vector-valued immediate reward function, with $n$ being the number of objectives of the problem. In single-objective RL, $n=1$ while in \emph{multi-objective reinforcement learning (MORL)}, $n>1$.

When $n=1$, the goal is to find the policy $\pi^*$ that maximizes the expected sum of discounted rewards:
\begin{equation}
\pi^*= \arg\max_\pi \mathbb{E} \left [ \sum_{t=0}^h \gamma^t r_t |\ \pi, s_0 \right ]
\end{equation}
In contrast, when $n > 1$, this sum can lead to returns for which, without any additional information, there is no clear winner (e.g., we cannot decide which return is optimal between $(0, 10)$, $(5, 5)$ or $(10, 0)$). A solution with values for all objectives lower than another is said to be \emph{dominated}. A solution for which it is impossible to improve one of the objectives without hampering another (e.g., any of the solutions in the previous example) is said to be \emph{Pareto-efficient}. The set of all Pareto-efficient solutions --- i.e., the set of all best possible compromises --- is called the \emph{Pareto front}.

We can represent the decision maker preferences with a \emph{utility function} $u : \mathbb{R}^n \rightarrow \mathbb{R}$, which converts the multi-objective returns back to a single scalar signal. Unfortunately, in most settings, the utility function is not known in advance: it either varies in each individual instance of the problem, or it cannot be easily formalized by the decision maker. Thus, in this work we assume that $u$ is unknown but monotonically increasing, and we focus on learning the full set of non-dominated solutions represented by the Pareto front. In addition, we must learn multiple separate policies that can reach all of the non-dominated solutions, since without knowing $u$ we must assume that any solution could be optimal. After learning, the decision maker can select their preferred solution and execute the corresponding policy.

We now define these concepts more formally. A policy $\pi$ is said to Pareto-dominate another policy $\pi'$ if its expected return $\mathbf{V}^{\pi}$, also called V-value, is higher or equal \emph{across all objectives} than $\mathbf{V}^{\pi'}$, and there exist at least an objective where $\mathbf{V}^{\pi}$ is better than $\mathbf{V}^{\pi'}$:
\begin{equation}
    \mathbf{V}^\pi \succ_P \mathbf{V}^{\pi'} \Longleftrightarrow (\forall i: \mathbf{V}^\pi_i \ge \mathbf{V}^{\pi'}_i) \land (\exists i: \mathbf{V}^\pi_i > \mathbf{V}^{\pi'}_i)
\end{equation}
where $\succ_P$ is the Pareto-dominance operator.

Our goal is then to find the set of policies that are not dominated by any other policy:
\begin{equation}
    \Pi^* = \{\pi \in \Pi\ |\ \nexists \pi' \in \Pi : \mathbf{V}^{\pi'} \succ_P \mathbf{V}^{\pi}\}
\end{equation}
where $\Pi$ is the set of all possible policies. $\Pi^*$ then maps to the Pareto front $\mathcal{F} = \{\mathbf{V}^{\pi}\ |\ \pi \in \Pi^* \}$. In general, we call any set of V-values mapped from a set of policies a \emph{solution set}, and any solution set composed only of non-dominated V-values a \emph{coverage set}. Thus, the Pareto front is the optimal coverage set of the problem.

Learning the full set of Pareto-efficient policies $\Pi^*$ requires that the policies $\pi^* \in \Pi^*$ are deterministic stationary policies \cite{roijers2013}. This is useful in settings where stochastic policies are not desired, such as the management of a hydroelectric power plant. In that scenario, the decision maker does not want to be presented with a policy that has a probability of completely draining the water reservoir even if that policy is optimal, as it would have catastrophic consequences for nearby towns \cite{hayes2021}.

\subsection{Multi-Objective Metrics}

Comparing the learned coverage sets of different algorithms is a non-trivial task, as one algorithm's output might dominate the other in some part of the objective-space, but be dominated in another. Intuitively, one would generally prefer the algorithm that obtains better returns for a wider range of utility functions. In this work we use several metrics to evaluate an algorithm's performance.

The most widely used metric in the literature is called the \emph{hypervolume} \cite{zitzler1999}. This metric evaluates the learned coverage set by computing its volume w.r.t. a fixed specified reference point. The reference point is taken as a lower bound on the achievable returns so that the volumes are always positive. Thus, the hypervolume envelops all possible V-values that are dominated by that coverage set, with more dominating coverage sets having a larger hypervolume. This metric is by definition the highest for the Pareto front, as no other possible solution can increase its volume (since they are all dominated). While the hypervolume metric is widely used and does give a measure of the coverage of a solution set, it can be difficult to interpret. The benefit of a certain increase or decrease in hypervolume is not readily apparent to the end user, and does not necessarily correlate to significant changes in expected utility. When working in high-dimensional objective-spaces, adding or removing a single point can lead to wildly different hypervolume values, especially if the point lies close to an extremum of the space.

To combat these limitations, we additionally evaluate our work with a different metric called the $\varepsilon$-indicator $I_\varepsilon$ \cite{zitzler2003}. $I_\varepsilon$ measures how close a coverage set is to the Pareto front $\mathcal{F}$. The $\varepsilon$-indicator of a coverage set $\hat{\Pi}$ is computed such that, for every $\mathbf{V}^{\pi}$ of the Pareto front, there exist a V-value in the coverage set that is at most $\epsilon$ smaller than $\mathbf{V}^{\pi}$:

\begin{equation}
\label{eq:epsilon-metric}
    I_{\varepsilon} = 
    \inf_{\varepsilon\in\mathbb{R}} 
    \{ 
    \forall \mathbf{V}^\pi\!{\in}\ \mathcal{F}, ~
    \exists \mathbf{V}^{\pi'}\!{\in}\ \hat{\Pi} :\ 
    || V^\pi - V_o^{\pi'} ||_{\infty} \le \varepsilon
    \}
\end{equation}

From a user's perspective, $I_\varepsilon$ has the intuitive meaning of showing that the proposed coverage set is at most $\varepsilon$ worse than any V-value of the Pareto front. The main disadvantage of this metric is that to compute it we need the true Pareto front, so it can only be used in test problems with a known structure.

While the $\varepsilon$-indicator has nice theoretical properties guaranteeing performance in the worst case \cite{zintgraf2015} --- i.e., the maximum utility loss (MUL) with respect to any possible utility function --- it does not give any information about the expected utility loss (EUL). This makes it a highly pessimistic metric; the $\varepsilon$-indicator will still report bad performance even if nearly the whole Pareto front is learned exactly, as long as a single point is not correctly modeled. 

While measuring the expected utility loss precisely requires knowing in advance the distribution of possible utility functions, we propose a variation on the $I_{\varepsilon}$ metric that aims to approximate an upper bound to this value, by making an additional assumption. 

In particular, we propose computing $I_{\varepsilon-mean}$, that assumes that each point in the Pareto front is equally likely to be selected as the best choice by a randomly sampled utility function. We compute $I_{\varepsilon-mean}$ by taking the mean of the computed $\varepsilon$ values, which are computed in the same way as the original $I_\varepsilon$ metric. Because the maximal utility loss of incorrectly preferring another vector in the coverage set over a given vector in the Pareto front is $\varepsilon\sqrt{n}L$ \cite{zintgraf2015}, where $L$ is the Lipschitz constant that described the level of continuity of the utility function, the $I_{\varepsilon-mean}$ metric describes an upper bound on the EUL, given that each vector in the Pareto set is (approximately) equally likely to be the vector that maximises the user's utility. 

\begin{figure}[t!]
    \centering
    \begin{tikzpicture}
      \begin{axis}[legend pos=south west, xmin=-1, xmax=5, ymin=-1, ymax=5,ytick={0,1,2,3,4,5}]
      \addplot[only marks,mark=x,mark size=3] coordinates {(-0.5,0)};
      \addplot[only marks,mark=*] coordinates {(1,4)(2,2)(4,1)};
      \addplot[only marks,mark=o] coordinates {(0.5,3)(0.75,2.3)(2.3,1)(3.3,0.7)};
      \fill [cyan,opacity=0.2] (-0.5,0) -- (-0.5,3) -- (0.5,3) -- (0.5,2.3) -- (0.75,2.3) --
      (0.75,1) -- (2.3,1) -- (2.3,0.7) -- (3.3,0.7) -- (3.3,0) -- cycle;
      \draw [dashed, -|] (0.5, 3) -- node[right]{$\varepsilon_1$} (0.5,4);
      \draw [dashed, -|] (2.3,1) -- node[right]{$\varepsilon_2$} (2.3,2);
      \draw [dashed, -|] (3.3,0.7) -- node[below]{$\varepsilon_3$} (4,0.7);
      \draw [opacity=0.2,line width=2,orange, dotted] (-0.5,0) -- (-0.5,4) -- (1,4) -- (1,2) --
      (2,2) -- (2,1) -- (4,1) -- (4,0) -- (-0.5,0);
      \end{axis}
    \end{tikzpicture}
    
    \caption{Pareto front and coverage set in 2-objective environment. The Pareto front is the set containing the best compromises that can be achieved (black dots). We want to learn a coverage set that is as close as possible to the true Pareto front (white dots). The hypervolume metric, in light blue, measures the volume of all dominated solutions w.r.t. some reference point (cross). The $\varepsilon$ metrics first compute the maximum distance between each point in the Pareto front and its closest point in the coverage set ($\varepsilon_i$). We can then take their maximum value to compute the $I_\varepsilon$ metric, or their mean value to obtain the $I_{\varepsilon-mean}$ metric of the coverage set.}
    \Description{The picture contains a set of points in a 2D space. The points belonging to the true Pareto front are black, while the other ones are drawn in black outlines. A light blue shade denotes the hypervolume of the coverage set.}
    \label{fig:paretofrontmetrics}
\end{figure}
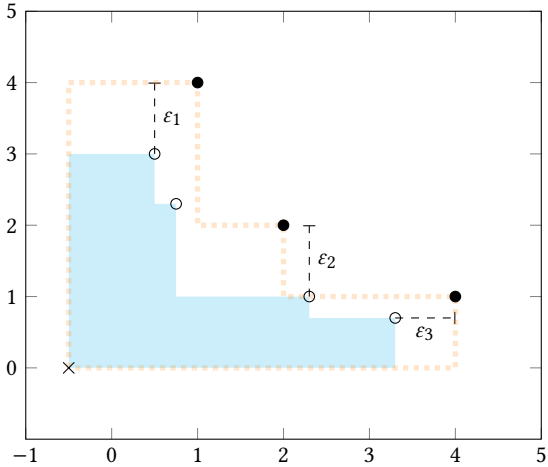

Figure \ref{fig:paretofrontmetrics} shows a visual representation of the hypervolume and $\varepsilon$ metrics in 2 dimensions.

\subsection{Reward Conditioned Policies}
\label{sec:rcp}

Our work is inspired by the Reward Conditioned Policies algorithm proposed by \cite{kumar2019,schmidhuber2019}.
Using neural networks as function approximators in RL comes with many challenges. One of them is that the target (e.g., the optimal action of the policy) is not known in advance --- as opposed to classical supervised learning where the ground-truth target is provided. As the behavior of the agent improves over time, the action used as target can change, often leading to hard-to-tune and brittle learners \cite{mnih2015,fu2019}. 

Instead of trying to continuously improve the policy by learning actions that should lead to the highest cumulative reward, Reward Conditioned Policies flips the problem, by learning actions that should lead to any \emph{desired} cumulative reward (be it high or low). In this way, all past trajectories can be reused for supervision, since their returns are known, as well as the actions needed to reach said returns. We can thus train a policy that, conditioned on a desired return, provides the optimal action to reach said return. By leveraging the generalization properties of neural networks, we can accumulate incrementally better experience by conditioning on increasingly higher reward-goals.

In our work, we condition our policy on a multi-objective return, such that we can execute policies to reach diverse points on the Pareto front. We propose a training regimen focused on increasing the current solution set uniformly across the whole objective-space to avoiding catastrophic forgetting.

\section{Related work}
\label{sec:related-work}

Current  MORL methods can be broadly divided into two main categories: \emph{single-policy} and \emph{multi-policy} algorithms \cite{vamplew2011}. In the first case, one tries to learn a single, optimal policy for a given set of preferences, i.e., for a known utility function. In the second case, the utility function is unknown (or uncertain) and the goal is to learn a set of policies that cover all possible utility functions, i.e., a coverage set.

Most of the recent work on MORL falls into the second category and assumes an unknown, but linear utility function. However, this restricts the solution set that can be learnt, as linear scalarization assumes the Pareto front to be convex. In the linear setting, the goal is to train an agent such that the optimal policy can be recovered for any preference weights. \cite{roijers2015} propose Optimistic Linear Support (OLS), a generic method that iteratively selects different sets of weights and calls a single-objective learner as subroutine to find the corresponding optimal policy. \cite{mossalam16} extend this method for Deep RL. Another approach, taken by \cite{barrett2008,hiraoka2009}, is to directly optimise on the coverage set without single-objective subroutine, by modifying the Bellman equation. Contrary to PCN, these methods are unable to discover any V-values on the concave regions of the Pareto front \cite{das1997}.

Using conditioned networks has been explored in MORL, but again restricted to the linear scalarization setting. In \cite{castelletti2012}, Fitted Q-Iteration (FQI) is extended to use a modified Q-network conditioned on \emph{preference weights} instead of target returns. Similarly, \cite{abels2019} use such a conditioned Q-network to extend Deep Q-Networks (DQN). Moreover, a similar network is used in \cite{runzhe2019}, in combination with a multi-objective Bellman operator.

Our work also can be related to imitation learning, as it also uses supervised learning to learn a policy \cite{sun2018,osa2018}. However, imitation learning requires expert trajectories to train on, while PCN generates its own set of trajectories.

When the utility function can be any monotonically-increasing function, \cite{vanmoffaert2014} adapt tabular Q-learning to directly learn the Pareto front. However, it is limited to discrete low-dimensional state-spaces.

Finally, \cite{parisi2014} learn to reach the Pareto front using a modified policy gradient search, and \cite{parisi2017} do this by modifying evolution strategies. Both algorithms make the same assumptions on the utility functions as PCN and are used as baselines in our experimental section (Section~\ref{sec:experiments}).

\section{Pareto Conditioned Networks}
\label{sec:method}

In this Section we introduce our main contribution, the Pareto Conditioned Networks (PCN) algorithm. The key idea behind our approach is to use supervised learning techniques to improve the policy instead of resorting to temporal-difference learning. As explained in Section~\ref{sec:rcp}, this eliminates the moving-target problem, resulting in stable learning.

PCN uses a single neural network that takes a tuple $\langle s, \hat{h}, \mathbf{\hat{R}}  \rangle$ as input. They represent, for state $s$, the return $\mathbf{\hat{R}}$ that PCN should reach at the end of the episode, i.e. the \emph{desired return} of the decision maker. The \emph{desired horizon} $\hat{h}$, that says how many timesteps should be executed before reaching $\mathbf{\hat{R}}$. At execution time, both $\hat{h}$ and $\mathbf{\hat{R}}$ are chosen by the decision maker at the start of the episode.

PCN's neural network has a separate output for each action $a_i \in \mathcal{A}$. Each output represents the confidence the network has that, by taking the corresponding action, the desired return will be reached in the desired number of timesteps. We can draw an analogy with a classification problem where the network should learn to classify $(s, \hat{h}, \mathbf{\hat{R}})$ to its corresponding label $a_i$.

Similarly as with classification, PCN requires a labeled dataset with training examples to learn a mapping from input to label. Contrary to classification, however, the data present in the dataset is not fixed. PCN collects data from the trajectories experienced while exploring the environment (see Section~\ref{sec:dataset}). Thus, the dataset improves over time, as we collect better and better trajectories. In particular, since the ability of PCN to reach Pareto-dominating solutions depends on the data on which its network is trained, we want to keep only relevant experiences. We do so by limiting the size of the dataset and pruning it in such a way so to keep only tuples with $\mathbf{\hat{R}}$'s from different parts of the objective-space (see Section~\ref{sec:updating-dataset}).

We collect new data for several episodes, after which we re-train the network with a number of batch updates from the new dataset (see Section~\ref{sec:training-policy}). This improves the policies induced by the network, which in turn allows to gather better data for the next training batch.

\subsection{Building the dataset}
\label{sec:dataset}

As mentioned before, each datapoint in PCN's dataset is comprised of an input $\langle s, \hat{h}, \mathbf{\hat{R}} \rangle$ and an output $a$. These are computed from observed transitions in the environment. As the dataset is empty at first, we execute a random policy on the environment for the first few episodes in order to collect a variety of trajectories.

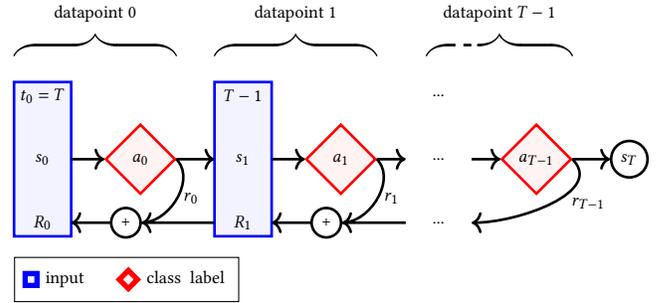
\begin{figure}[t]
    \centering
    \begin{tikzpicture}[scale=0.65,every node/.style={scale=0.75}]
\node [] at (0,1.3) (t0) {$t_0 = T$};
\node [] at (0,0)  (s0) {$s_0$};
\node [] at (0,-1.3)  (r0) {$R_0$};
\node at (2,0) (a0) {$a_0$};
\node [draw=red,diamond,line width=1,fill=red,fill opacity=0.05] at (2,0) (a0box) {\phantom{$a_{T-1}$}};

\draw [blue,line width=1,fill=blue,fill opacity=0.05] (t0.north west) rectangle (r0.south east -| t0.north east);
\draw [->,line width=1] (s0-|t0.east) -- (a0box);

\node [] at (4.1,1.3) (t1) {$T - 1$};
\node [] at (4.1,1.3) (t1box) {\phantom{$t_0 = T$}};
\node [] at (4.1,0)  (s1) {$s_1$};
\node [] at (4.1,-1.3)  (r1) {$R_1$};
\node at (6.1,0) (a1) {$a_1$};
\node [draw=red,diamond,line width=1,fill=red,fill opacity=0.05] at (6.1,0) (a1box) {\phantom{$a_{T-1}$}};

\draw [blue,line width=1,fill=blue,fill opacity=0.05] (t1box.north west) rectangle (r1.south east -| t1box.north east);

\draw [->,line width=1] (a0box) -- (s1-|t1box.west);
\draw [->,line width=1] (s1-|t1box.east) -- (a1box);

\node [] at (8.1,1.3)  (tn) {...};
\node [] at (8.1,0)  (sn) {...};
\node [] at (8.1,0)  (snbox) {\phantom{$t_0 = T a$}};
\node [] at (8.1,-1.3) (rn) {...};

\draw [->,line width=1] (a1box) -- (snbox.west);

\node [draw=red,diamond,line width=1,fill=red,fill opacity=0.05,text opacity=1] at (10.1,0) (at) {$a_{T-1}$};

\draw [->,line width=1] (snbox.east) -- (at);

\node [draw,circle,line width=1] at (12,0)  (st) {$s_T$};

\draw [->,line width=1] (at) -- (st);

\node [draw,circle,line width=1] at (a0.west|-0,-1.3) (p0) {+};
\node [draw,circle,line width=1] at (a1.west|-0,-1.3) (p1) {+};

\draw [line width=1] (a0box.east) to [out=-45,in=0] node[right] {$r_0$} (p0.east);
\draw [line width=1] (a1box.east) to [out=-45,in=0] node[right] {$r_1$} (p1.east);
\draw [->,line width=1] (at.east) to [out=-45,in=0] node[right=5pt] {$r_{T-1}$}(snbox.east |- rn);

\draw [->,line width=1] (p0) -- (t0.east |- r0);
\draw [->,line width=1] (t1box.west |- r1) -- (p0);
\draw [->,line width=1] (p1) -- (t1box.east |- r1);
\draw [->,line width=1] (snbox.west |- rn) -- (p1);

\draw [decoration={calligraphic brace,amplitude=8pt,raise=10pt},line width=1,decorate]
    (t0.north west) -- node[above=20pt]{datapoint $0$}(a0box.east|-t0.north);
\draw [decoration={calligraphic brace,amplitude=8pt,raise=10pt},line width=1,decorate]
    (t1box.north west) -- node[above=20pt]{datapoint $1$}(a1box.east|-t1box.north);
\draw [pen colour={white},decoration={calligraphic brace,amplitude=8pt,raise=10pt},line width=1,decorate]
    (tn.north west|-t0.north) -- node[above=20pt]{datapoint $T-1$}(at.east|-t0.north);
\begin{scope}
\clip (at.west |- 0,3) rectangle (14,0);
\draw [decoration={calligraphic brace,amplitude=8pt,raise=10pt},line width=1,decorate]
    (tn.west|-t0.north) -- (at.east|-t0.north);
\end{scope}

\begin{scope}
\clip (r0.south) rectangle (at.west |- 0,3);
\draw [decoration={calligraphic brace,amplitude=8pt,raise=10pt},line width=1,dashed,decorate]
    (tn.west|-t0.north) -- (at.east|-t0.north);
\end{scope}

\coordinate (legend) at (t0.west |- 0,-2);

\matrix [matrix of nodes,draw,below right,column sep=10pt] at (legend) {
  \node [shape=rectangle,line width=2,draw=blue,fill=blue,fill opacity=0.05,label=right:input] {}; &
  \node [shape=diamond,inner sep=2.5pt,line width=2,draw=red,fill=red,fill
  opacity=0.05,label=right:class\phantom{p}label] {}; \\
};

\end{tikzpicture}
    \caption{Conversion from a trajectory to labeled datapoints. For each timestep, we extract a single datapoint. The input (blue) is composed of the state at that timestep, and total return and number of timesteps until the end of the episode. The label (red) is the action taken at that timestep.}
    \Description{A series of boxes and diamonds connected by arrows, showing the progression of state, returns, horizons and actions for each timestep along a single trajectory.}
    \label{fig:dataset}
\end{figure}

After each episode is completed we store its trajectory. Then, for each timestep $t$ of the trajectory, we know how many timesteps are left until the end is reached, i.e., the episode's horizon $h_t=T-t$. We can also compute the cumulative reward obtained from timestep $t$ onward, i.e., $\mathbf{R}_t = \Sigma_{i=t}^{T}{\gamma^i \mathbf{r}_i}$. Since for this trajectory executing action $a_t$ in state $s_t$ resulted in return $\mathbf{R}_t$ in $h_t$ timesteps, we add a datapoint with input $\langle s, \hat{h}, \mathbf{\hat{R}}\rangle = \langle s_t, h_t, \mathbf{R}_t \rangle$ and output $a=a_t$ to the dataset. In other words, when the observed return corresponds to the desired return in that state, then $a_t$ is the optimal action to take. Figure~\ref{fig:dataset} shows how a full trajectory is decomposed into individual datapoints.

\subsection{Training the Network}
\label{sec:training-policy}

PCN's network architecture uses a separate embedding for the state and another one for the desired return and horizon. The desired return and horizon are concatenated together and multiplied by a scaling factor to normalize their values. They then pass through a single fully connected layer followed by a sigmoid function. Similarly, the state-embedding also ends with a fully connected layer and sigmoid activation. Both layers have the same number of output nodes (we use 64 for all our experiments), which are combined together using the Hadamard product. The Hadamard product has been shown to make a more effective use of conditioning variables \cite{perez2018film}, and the sigmoid used on both outputs ensure that both embeddings are equally important. Finally, the resulting output passes through a multilayer perceptron that has a separate output node for each action, and a single hidden layer of 64 nodes with a ReLU activation function. 

PCN trains the network as a classification problem, where each class represents a different action. Transitions $x= \langle s_t, h_t, \mathbf{R}_t \rangle, y=a_t$ are sampled from the dataset, and the ground-truth output $y$ is compared with the predicted output $\hat{y}=\pi(s_t, h_t, R_t)$. The predictor (i.e., the policy) is then updated using the cross-entropy loss function:
\begin{equation}
    H = -\sum_{a \in \mathcal{A}}{y_a \log \pi(a|s_t,h_t,\mathbf{R}_t)}
\end{equation}
where $y_a = 1$ if $a = a_t$ and $y_a = 0$ otherwise.

The network is re-trained periodically, but only after a set number of episodes (which is a hyperparameter that depends on the problem), to ensure that enough new experience has been collected and that the underlying dataset has been improved sufficiently.

\subsection{Policy Exploration}
\label{sec:selecting-target}

As our dataset is composed of transitions collected from training experience, we can see that the quality of our dataset crucially depends on the quality of the executed trajectories. It is unrealistic to expect PCN to reliably produce trajectories with high-valued desired returns when it has only been trained on datapoints originating from random trajectories. Rather, we can expect PCN to produce trajectories with returns in the range of the ones from the current training data. Therefore, if we obtain trajectories reaching high returns, PCN will be able to confidently return high-return policies.

PCN leverages the fact that, due to the generalization capabilities of neural networks, the policies obtained from the network will still be reliable even if the desired return is marginally higher than what is present in the training data. In fact, they will perform similar actions to those in the training data, but lead to a higher return. Thus, we incrementally condition the network on better and better returns, in order to obtain trajectories that extend the boundaries of PCN's current coverage set.

More precisely, we randomly select a non-dominated return $\mathbf{R}_{nd}$ and its corresponding horizon $\hat{h}$ from the dataset. By randomly picking a non-dominated return from the entire coverage set we ensure equal chance of improvement to each part of the objective space. However, using $\mathbf{R}_{nd}$ exactly would induce the network to only replicate the already observed sampled trajectory so, as a second step, we choose a single objective $o$ to improve upon. We then increase the desired value for that objective to obtain a new target return. PCN determines the magnitude of the increase by computing the standard deviation $\sigma_o$ for the selected objective, using all non-dominated returns from the trajectories in the dataset. The magnitude is then sampled from the uniform distribution $U(0, \sigma_o)$ and added to $R_{nd,o}$ to form our desired return $\mathbf{\hat{R}}$. By restricting the improvement to at most $\sigma_o$, $\mathbf{\hat{R}}$ stays in the range of possible achievable returns and, by only modifying one objective at a time, the changes to the network's input compared to the training data are kept at a minimum.

With $\mathbf{\hat{R}}$ and $\hat{h}$ selected, PCN can condition its network and act during the training episode. At the start of the episode, $\langle s_0, \hat{h}, \mathbf{\hat{R}}\rangle$ results in executing $a_0$ and observing $\mathbf{r}_0, s_1$. PCN then updates the desired return and horizon such that they stay consistent throughout the episode: $\mathbf{\hat{R}} \longleftarrow \mathbf{\hat{R}} - \mathbf{r}_0$ and $\hat{h} \longleftarrow \max(\hat{h}-1, 1)$. We ensure that $\hat{h}$ is at least 1 to avoid impossible desired horizons. \method can then choose an action for $s_1$. This process is repeated until PCN encounters a terminal state. The trajectory is then added to the dataset, using the conversion to datapoints explained in Section~\ref{sec:dataset}.

To increase the range of observations in the environment during training, PCN samples actions from a categorical distribution with each action's confidence score corresponding to its probability of being sampled. This is done  by using a softmax function on the network's output. Please note that at execution time --- i.e., after the training process --- we use a deterministic policy by systematically selecting the action with the highest confidence. This is because, as mentioned in Section~\ref{sec:background}, only deterministic policies are optimal when learning the complete set of Pareto-efficient policies. 

\begin{figure}[t]
    \centering
    \begin{tikzpicture}[
        bluenode/.style={shape=circle, draw=blue, fill=blue, inner sep=2pt, outer sep=4pt},
        rednode/.style={shape=circle, draw=red, fill=red, inner sep=2pt, outer sep=4pt}
      ]
      \begin{axis}[legend pos=south west, xmin=0, xmax=5, ymin=0, ymax=5,xtick={1,2,3,4,5},ytick={1,2,3,4,5}]
      \addplot[only marks,mark=o] coordinates {(1,1)(1.6,2.8)(2.5,3)(2.9,2)};
      \addplot[only marks,mark=o] coordinates {(2.7,2.9)};
      \addplot[only marks,mark=*,orange] coordinates {(2,4)(3,3)(4,1)};
      \addplot[only marks,mark=*,black] coordinates {(1.9,3.5)};
      \draw [dashed,blue] (1.6,2.8) -- (1.6,0);
      \draw [dashed,blue] (2,4) -- (2,0);
      \draw [dashed,blue] (2.5,3) -- (0,3);
      \draw [dashed,blue] (2,4) -- (0,4);
      \draw [blue,line width=3pt] (1.6,0) -- (2,0);
      \draw [blue,line width=3pt] (0,3) -- (0,4);
      \draw [red,line width=0.7pt,-latex] (1.9,3.5) -- (1.985,3.94); 
      \end{axis}
      \matrix [draw,below left=1cm] at (current bounding box.north east) {
        \node [rednode,label=right:$I_{l2}$] {}; \\
        \node [bluenode,label=right:$I_{cd}$] {}; \\
      };
    \end{tikzpicture}
    \caption{The $I_{ds}$ metric for the solid black dot combines its negative L2-norm distance (red arrow) to its closest non-dominated neighbor (orange), and its crowding distance as the sum, for each dimension, of the max distance between a point's upper and lower neighbour (blue).}
    \Description{A 2D plot showing a current solution set, with both dominated and non-dominated points. An arrow denotes the L2-norm distance from a point to its closest non-dominated point. A set of blue dashed lines denote the crowding distance, as distances of that same points to its closest neighbors in both dimensions.}
    \label{fig:metric}
\end{figure}
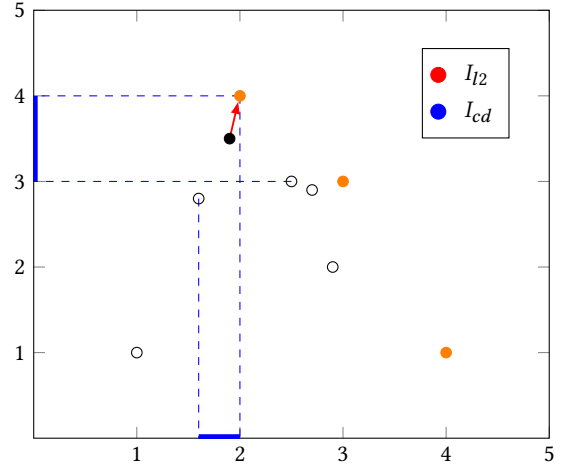

\subsection{Updating the Dataset}
\label{sec:updating-dataset}

As it collects new experience, PCN needs to use it to train its network without forgetting about previous, relevant experience. Unfortunately, measuring relevance of data in our setting is non-trivial. We mainly care about the non-dominated solutions, since those are the ones that compose our coverage set, so ideally we would only keep their associated trajectories in our dataset. However, focusing solely on the current coverage set can lead to performance degradation if it is composed of too few V-values. This is because, during exploration (see Section \ref{sec:selecting-target}), we will keep collecting very similar trajectories to the ones we already have. In turn, this will reinforce PCN's strategy to focus only on these few policies, disrupting the learning process. Thus, throughout training we keep a dataset of $N$ trajectories, favoring trajectories that span different parts of the objective-space while removing highly clustered solutions. To do this we employ an additional metric, the \emph{crowding distance}, that assigns a lower score to points with close neighbors. We then combine both the distance to the coverage set and the crowding distance in a single metric, which we use to prune less relevant points from the dataset.

We measure our preference for non-dominated V-values by computing for each solution its negative L2-norm distance, $I_{l2}$, to its closest non-dominated solution in the dataset. Non-dominated V-values thus score the highest with $I_{l2} = 0$.

\begin{equation}
    I_{l2, i} = -\min \norm{p_{i}-p_{j}}_2, p_{j} \in \hat{\Pi}
\end{equation}
where $i$ is the index of the $i$-th solution in the dataset and $p_j$ is a non-dominated point in the current coverage set $\hat{\Pi}$.

We measure the level of clustering of a V-value, $I_{cd}$, using the crowding distance \cite{deb2000}. It assigns a score for each solution based on the distance between its neighbours in each dimension. Thus, V-values with close neighbors will have lower scores, while more isolated points will have higher scores. Algorithm \ref{algo_crowding} shows how the crowding distance is computed in practice.

\begin{algorithm}
\caption{Crowding Distance}
\label{algo_crowding}
\begin{algorithmic}[1]
    \Require{$p$ points}
    \Ensure{The crowding distance $I_{cd}$}
    \For{$o \gets 0$ \textbf{to} $n$}
        \State $s_i \gets \argsort p_{.,o}$
        \For{$j \gets 0$ \textbf{to} $\texttt{len}(p)$}
            \State $n_u, n_d \gets s_{i_{j+1}}, s_{i_{j-1}}$
            \State $c_{s_{i_j}, o} = p_{n_u} - p_{n_d}$
        \EndFor
    \EndFor
    \State \Return $I_{cd} \gets \sum_{o}{c_{.,o}}$
\end{algorithmic}
\end{algorithm}

We then combine $I_{l2}$ with $I_{cd}$ in a single score metric, which we call \emph{dominating score}, $I_{ds}$. We define it as:

\begin{equation}
    I_{ds,i} =
    \begin{cases}
        I_{l2,i} & \texttt{if } I_{cd,i} > 0.2 \\
        2(I_{l2,i}-c) & \texttt{if } I_{cd,i} \leq 0.2
    \end{cases}
\end{equation}
Note that since $I_{l2}$ corresponds to a negative distance, if a point is crowded --- i.e., $I_{cd,i} \leq 0.2$ -- we double the distance penalty. In addition, we add an additional small penalty $c$ to crowded points to prune, to detect and prune duplicate points on the coverage set. Figure~\ref{fig:metric} shows an example on how to compute the L2-norm and crowding distances for a given solution set.

\begin{table*}[ht!]
    \centering
    
\begin{tabular}{|c|c|c|c||c|c|c|}
     \hline
     & \multicolumn{3}{c||}{hypervolume} & \multicolumn{3}{c|}{$I_\varepsilon$ indicator} \\
      \hhline{|=|= = =||= = =|}
      & \method (ours) & MONES & RA & \method (ours) & MONES & RA\\
      \hline
     DST & $\mathbf{22845.40 \pm 19.20^*}$ & $17384.83 \pm 6521.10$ & $22437.40 \pm 49.20$ & $\mathbf{0.039 \pm 0.087^*}$ & $0.687 \pm 0.222$ & $0.667 \pm 0.000$\\
     Minecart & $\mathbf{197.56 \pm 0.70^*}$ & $123.81 \pm 23.03$ & $123.92 \pm 0.25$ & $\mathbf{0.271 \pm 0.087^*}$ & $1.596 \pm 0.889$ & $1.000 \pm 0.000$\\
     Crossroad & $\mathbf{539.53 \pm 6.27^*}$ & $429.09 \pm 27.47$ & $466.02 \pm 31.23$ & $\mathbf{0.247 \pm 0.172^*}$ & $0.660 \pm 0.200$ & $0.408 \pm 0.039$ \\
     \hline
     
\end{tabular}
    \caption{Mean and standard deviation of hypervolume and $\varepsilon$-indicator across all 5 runs for all algorithms. For hypervolume, higher is better. For $I_\varepsilon$, lower is better. Best results are highlighted in bold and with an asterisk.}
    \label{tab:hypervolume}
\end{table*}

\section{Experiments}
\label{sec:experiments}

Most state-of-the-art algorithms in MORL assume that the utility function can be expressed as a linear scalarization. This makes them unsuitable as relevant baselines, as their setting is different from ours. Thus, for all experiments, we compare PCN with 2 baselines that share our same assumptions on the utility function and that learn a set of policies that estimates the whole Pareto front. The first baseline, Multi-Objective Natural Evolution Strategies (MONES) \cite{parisi2017} uses a parametrized policy. It learns a distribution over the policy parameters such that sampling from this distribution produces a Pareto-efficient policy. Different samples produce different Pareto-efficient policies, each leading to a different non-dominated return. The advantage of this method is that it can produce an infinite number of different policies. However, the main drawback is that we do not know the return of the sampled policy without executing it first.

The second baseline used is the Radial Algorithm (RA) \cite{parisi2014}. RA trains a fixed number of independent policies. Each policy is trained using a policy gradient algorithm, where the gradients w.r.t. the different objectives are weighted together. Using different weights on these gradients produces distinct policies, each aiming for different regions of the objective-space. The main disadvantage of this approach is that every new policy is learnt independently, disregarding potentially useful experience encountered by other policies.

In contrast, our approach makes efficient usage of encountered experience as it learns a single network that, when conditioned on a desired return, produces a policy with predictable behavior.

When not mentioned otherwise, all results are averaged over 5 runs.

The results for all experiments are summarized in Table~\ref{tab:hypervolume}. Figures~\ref{fig:dst}-\ref{fig:sumo} show, for each environment, the coverage set found by each algorithm, with dominated solutions filtered out.

We experimentally validate our method on three multi-objective benchmarks:
\begin{itemize}
\item Deep-Sea-Treasure \cite{vamplew2011}, a well-known 2-objective grid-world problem (see Section~\ref{sec:dst}),
\item Minecart \cite{abels2019}, a 3-objective problem with a continuous state-space (see Section~\ref{sec:mine}),
\item Crossroad, a novel 2-objective traffic environment, with high-dimensional pixel-like states (see Section~\ref{sec:cross}).
\end{itemize}

In addition, in Section~\ref{sec:scaling} we propose an additional novel high-objective environment, Walkroom, where we test our algorithms with up to 9 objectives.

\subsection{Deep-Sea-Treasure}
\label{sec:dst}

Deep-Sea-Treasure (DST) is a well known environment in multi-objective literature \cite{vamplew2011}, where the agent controls a submarine in search for treasure hidden in the depth of the ocean. The agent must balance a trade-off between fuel consumption and treasure value. Navigating consumes fuel, but deeper treasures are worth more than shallow ones.

DST is a fairly small environment, which allows us to compute Pareto front analytically. It is composed of 10 different points, which form a concave front.

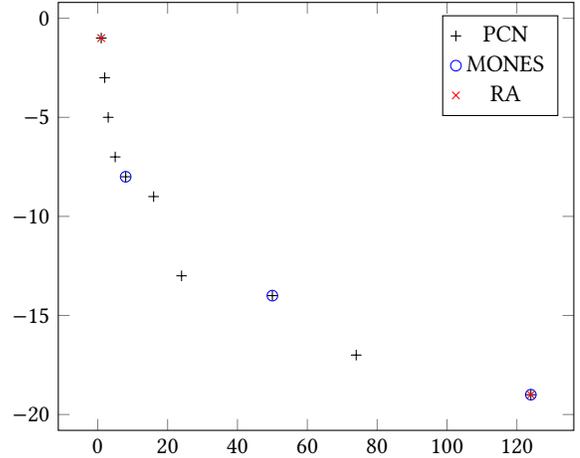
\begin{figure}
    \centering
    \begin{tikzpicture}
      \begin{axis}[legend pos=north east]
      \addplot[only marks,mark=+] table [x=o0, y=o1, col sep=comma] {data/dst/udrl.csv};
      \addplot[only marks,mark=o,blue] table [x=o0, y=o1, col sep=comma] {data/dst/mones.csv};
      \addplot[only marks,mark=x,red] table [x=o0, y=o1, col sep=comma] {data/dst/ra.csv};
      \addlegendentry{PCN}
      \addlegendentry{MONES}
      \addlegendentry{RA}
      \end{axis}
    \end{tikzpicture}
    
    \caption{Best coverage sets for each algorithm in the Deep Sea Treasure environment. PCN is the only algorithm that recovered the full Pareto front.}
    \Description{A 2D plot showing dots for each solution found by the algorithms in the Deep Sea Treasure environment.}
    \label{fig:dst}
\end{figure}

Figure~\ref{fig:dst} shows the discovered points of the Pareto front by PCN and our two baselines. Only PCN is able to fully recover the Pareto front. RA only discovers the extrema, and is unable to find any of the points in the concave part of the front. While MONES performs slightly better, the number and value of points it discovered was highly variable depending on the run, explaining the high variance seen in Table~\ref{tab:hypervolume}.

\subsection{Minecart}
\label{sec:mine}

Minecart is a complex environment with a continuous state-space \cite{abels2019}. Starting at a base station, the agent controls a cart with the goal to extract ores from mines scattered in the environment, and sell them back at its base.

The agent can execute 6 possible actions: it can accelerate, decelerate and rotate the cart to the left or right. It can mine ores, which will only be effective if it is located in a mine. It can also simply do nothing. There are 2 types of ores located in the different mines, where each type represents a separate objective. The cart has a limited capacity, so it cannot be filled indefinitely. Finally, actions consume fuel, making the Minecart problem a 3-objective problem.

\begin{figure}
    \centering
    \begin{tikzpicture}
      \begin{axis}[legend pos=north east,view={20}{70},zmin=-70,zmax=0]
      \pgfplotstableread[col sep=comma]{./data/minecart/mones-nd.csv}{\mones}
      \pgfplotstableread[col sep=comma]{./data/minecart/udrl.csv}{\udrl}
      \pgfplotstableread[col sep=comma]{./data/minecart/ra-nd.csv}{\ra}

      \addplot3[only marks,mark=+,mark options={line width=.6}] table [x=o0, y=o1, z=o2, col sep=comma] {\udrl};
      \addplot3[only marks,mark=o,blue,mark options={line width=.6}] table [x=o0, y=o1, z=o2, col sep=comma] {\mones};
      \addplot3[only marks,mark=x,red,mark options={line width=.5,scale=1.2}] table [x=o0, y=o1, z=o2, col sep=comma] {\ra};

      \addlegendentry{PCN}
      \addlegendentry{MONES}
      \addlegendentry{RA}

      \addplot3[only marks,mark=+,dashed,mark options={xscale=1.5, xslant=.37,yslant=-.3, dash
      pattern=on 1pt off .5pt, opacity=0.6}] table [x=o0, y=o1, z expr={-70.0}, col sep=comma] {\udrl};
      \addplot3[only marks,mark=o,blue,dashed,mark options={xscale=1.5, xslant=.37,yslant=-.3, dash
      pattern=on 1pt off 1pt, opacity=0.6}] table [x=o0, y=o1, z expr={-70.0}, col sep=comma] {\mones};
      \addplot3[only marks,mark=x,red,dashed,mark options={xscale=1.5, xslant=.37,yslant=-.3, dash
      pattern=on 1pt off .5pt,opacity=0.6}] table [x=o0, y=o1, z expr={-70}, col sep=comma] {\ra};

      \addplot3[
           draw=none,
           error bars/z dir=minus,
           error bars/z explicit,
           error bars/error mark=none,
           error bars/error bar style={
             dashed, dash pattern=on 1pt off 2pt on .3pt off 2pt,
             opacity=0.6
           },
       ] table [col sep=comma, x=o0, y=o1, z=o2, z error expr={\thisrow{o2}+70}] {\udrl};
      \addplot3[
           draw=none,
           error bars/z dir=minus,
           error bars/z explicit,
           error bars/error mark=none,
           error bars/error bar style={
             dashed, dash pattern=on 1pt off 2pt, dash phase=0pt,
             opacity=0.6
           },
           blue,
       ] table [col sep=comma, x=o0, y=o1, z=o2, z error expr={\thisrow{o2}+70}] {\mones};
      \addplot3[
           draw=none,
           error bars/z dir=minus,
           error bars/z explicit,
           error bars/error mark=none,
           error bars/error bar style={
             dashed, dash pattern=on .5pt off 1pt, dash phase=0pt,
             opacity=0.6
           },
           red,
       ] table [col sep=comma, x=o0, y=o1, z=o2, z error expr={\thisrow{o2}+70}] {\ra};

      \end{axis}
    \end{tikzpicture}
    
    \caption{Best coverage sets for each algorithm in the Minecart environment. The straight shape of PCN's coverage set is due to the cart weight limit. PCN's coverage set fully dominates the ones from the baselines.}
    \Description{A 3D plot showing dots for each solution found by the algorithms in the Minecart environment.}
    \label{fig:minecart}
\end{figure}
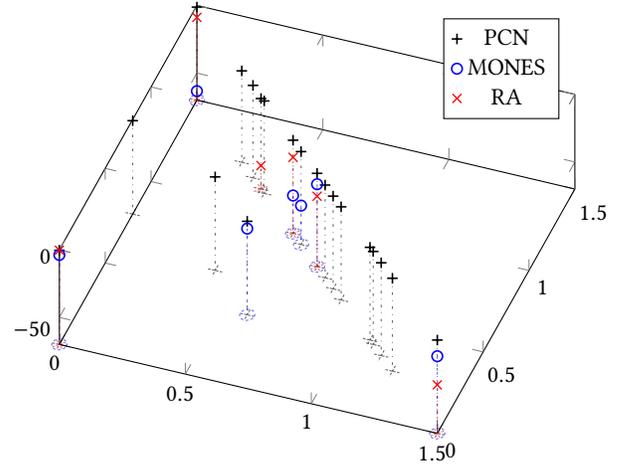

Because the cart has a limited capacity to store ores, the agent must decide the ratio of each ore present in the cart. This is why, as can be seen in Figure~\ref{fig:minecart}, the vast majority of policies discovered by PCN are laid out in a straight line: the sum of $R_0$ and $R_1$ equals $1.5$ (which is the cart capacity). A few policies only partially fill the cart, saving a bit of fuel in the process. In comparison, the coverage sets discovered by both baselines only contain a small subset of the possible ore ratios, and they systematically consume more fuel than our method. Finally, Table~\ref{tab:hypervolume} reports that PCN achieves a low variance in hypervolume across the different runs, which shows that PCN is consistent in finding these diverse and efficient policies.

\subsection{Crossroad}
\label{sec:cross}

We evaluate our proposed method on a novel traffic environment, Crossroad, developed using the SUMO framework \cite{SUMO2018}. In this environment, the agent controls the traffic lights at a busy intersection between two bidirectional roads, an horizontal one with two lanes and a vertical one with a single lane. The agent can choose between 2 actions, either switching the lights or not. We consider 2 objectives: the first is the traffic flow, computed as the number of cars that leave the intersection. The second objective is the car waiting time, i.e., the number of timesteps a car has to wait before exiting the crossroad. Thus, favoring traffic on the horizontal road will favor the first objective, while alternating often will favor the second.

\begin{figure}
    \centering
    \begin{tikzpicture}
      \begin{axis}[legend pos=south west]
      \addplot[only marks,mark=+] table [x=o0, y=o1, col sep=comma] {data/sumo/udrl.csv};
      \addplot[only marks,mark=o,blue] table [x=o0, y=o1, col sep=comma] {data/sumo/mones-nd.csv};
      \addplot[only marks,mark=x,red] table [x=o0, y=o1, col sep=comma] {data/sumo/ra-nd.csv};
      \addlegendentry{PCN}
      \addlegendentry{MONES}
      \addlegendentry{RA}
      \end{axis}
    \end{tikzpicture}
    
    \caption{Best coverage sets for each algorithm in the Crossroad environment. PCN fully dominates the baselines but for the naive policy that never switches the traffic lights (bottom-right point).}
    \Description{A 2D plot showing dots for each solution found by the algorithms in the Crossroad environment.}
    \label{fig:sumo}
\end{figure}
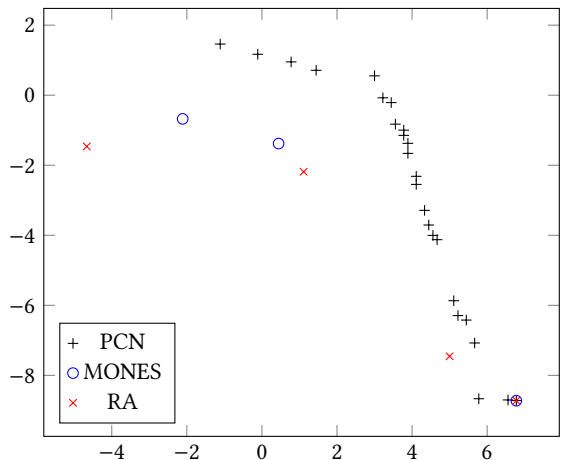

Figure~\ref{fig:sumo} shows the coverage sets found by PCN and the two baselines. The points discovered by our method dominate the ones found by the baselines across nearly the whole objective-space. There is a single exception is the rightmost extremum, which corresponds to the policy that never switches the light, only allowing cars from the major road to cross the intersection. This policy is pretty simple, as it consists of always performing the same action, which explains why all methods find it.

\section{Scaling Up the Objective-Space}
\label{sec:scaling}

Our experimental section shows that our proposed method significantly outperforms the baselines in several settings, with discrete, continuous and high-dimensional state-spaces. It also empirically shows that the coverage sets discovered by \method contain more non-dominated solutions than our selected baselines. Nevertheless, the vast majority of benchmarks used in MORL, including the ones used in our experiments, are limited to 2, sometimes 3 different objectives \cite{hayes2021}. 

To get a better understanding of the performance of \method w.r.t. the number of objectives, we devise a synthetic environment, \emph{Walkroom}, that can be instanced with an arbitrary number of objectives. Walkroom takes inspiration from Deep Sea Treasure, and is modeled as an $n$-dimensional grid-world in which the agent can move in every cardinal direction. Thus, the dimensions correspond to the number of objectives of the environment. The action space increases linearly with the number of objectives ($|\mathcal{A}| = 2n$). At each timestep, the agent receives a $-1$ reward for the objective corresponding to its moving dimension, and a $0$ reward for all other objectives. There are no other rewards. Similarly to Deep Sea Treasure, Walkroom has a set of goal states positioned along an uneven border, so that reaching each goal state results in a different Pareto-efficient solution. The optimal policies are thus to go directly towards any of the border-positions, at which point the episode ends.

We evaluate each method in a set of randomly generated Walkroom environments, from 2 to 9 dimensions. We perform 20 runs for every algorithm, on every version of the environment ($n=2,\dots,9$). Note that we do not plot the discovered coverage sets since this is not possible for $n > 3$. Instead, we show boxplots of the hypervolume and $\varepsilon$ metrics computed on all runs.  Since the Pareto fronts can be computed analytically when the environments are generated, the $\varepsilon$ metrics give us an accurate representation of the coverage of each solution set found by each learning algorithm.

\begin{figure}[t]
    \centering
    \includegraphics[width=\columnwidth]{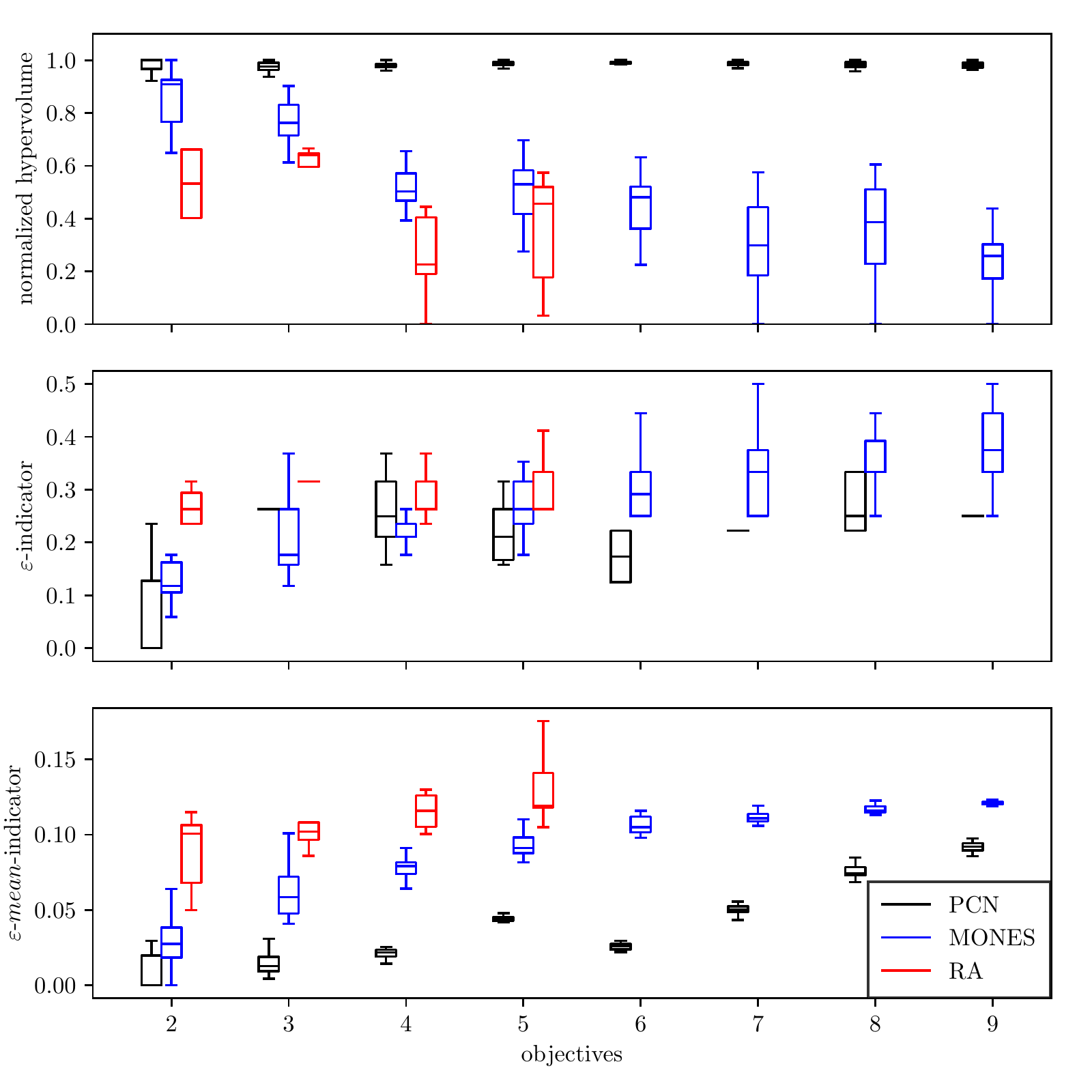}
    \caption{Box plots of normalized hypervolume, $I_{\varepsilon}$ and $I_{\varepsilon-mean}$ in the Walkroom environments, from 2 to 9 objectives. We normalized the hypervolumes as their true values scale exponentially with the number of objectives (for $n=9$, in the order of $10^{9}$). For hypervolume, higher is better. For $\varepsilon$-indicator metrics, lower is better.}
    \Description{Box plot of various metrics obtained in the Walkroom environments.}
    \label{fig:walkroom}
\end{figure}

Results are summarized in Figure~\ref{fig:walkroom}. RA performs poorly compared to the other algorithms across all metrics. This is because the number of policies that RA requires to cover the objective-space increases exponentially with the number of objectives. Thus, training time must be split between more and more policies, which reduces their individual performance. For the same reason, we were unfortunately unable to compare in RA in environments with more than 5 objectives, due to its exponential computational costs.

For $n \leq 4$, MONES achieves similar or better scores than \method in the $I_{\varepsilon}$ metric. This is likely due to \method missing some points from the Pareto front, which impact this metric significantly. However, we can see from the hypervolume and $I_{\varepsilon-mean}$ metrics that \method generally performs better than MONES on the rest of the Pareto front. Surprisingly, for $n > 4$ MONES performs significantly worse than \method. This might be because the parameter distributions of MONES are not able to explore efficiently in very large dimensional spaces. 

\section{Conclusion}

We have presented a novel algorithm, Pareto Conditioned Networks, which is able to efficiently and effectively learn coverage sets in multi-objective sequential problems. PCN uses a single neural network to generalize experience across all possible multi-objective returns, learning coverage sets even in concave Pareto fronts.

We evaluated the empirical performance of PCN in several environments against state-of-the-art benchmarks. PCN was able to consistently obtain higher returns than the baselines throughout the whole objective-space, demonstrating its ability to exhaustively discover optimal coverage sets. In addition, PCN demonstrated its ability to learn the Pareto front even when dealing with a large number of objectives.

While PCN can work in continuous state-spaces, its network architecture is currently limited to discrete action-spaces. However, the main ideas behind PCN do not change for continuous action-spaces and PCN can be extended to continuous actions by changing the classification problem to a regression one. We leave this for future work.



\begin{acks}
The authors would like to acknowledge FWO (Fonds Wetenschappelijk Onderzoek) for their support through the SB grant of Eugenio Bargiacchi (\#1SA2820N).
This research was additionally supported by funding from the Flemish Government under the “Onderzoeksprogramma Artificiële Intelligentie (AI) Vlaanderen” programme.

We would also like to thank Diederik M. Roijers for helpful feedback.
\end{acks}



\bibliographystyle{ACM-Reference-Format} 
\bibliography{references}

\end{document}